\documentclass[10pt,twocolumn,letterpaper]{article}
\usepackage{titling}

\usepackage[pagenumbers]{wacv}

\usepackage{graphicx}
\usepackage{amsmath}
\usepackage{amssymb}
\usepackage{booktabs}

\newcommand{\methodName}{HCCNet\xspace}

\usepackage{xr}
\makeatletter
\newcommand*{\addFileDependency}[1]{%
  \typeout{(#1)}
  \@addtofilelist{#1}
  \IfFileExists{#1}{}{\typeout{No file #1.}}
}
\makeatother

\usepackage{graphicx}
\usepackage{amsmath}
\usepackage{amssymb}
\usepackage{booktabs}
\usepackage{algorithm}
\usepackage{dsfont}
\usepackage{algorithm}
\usepackage{algpseudocode}
\usepackage{algorithmicx}
\usepackage{mathtools}
\usepackage{multirow}
\usepackage{graphics}
\usepackage{wrapfig}
\usepackage{bbm}
\usepackage{xspace}
\usepackage{xcolor}
\usepackage{tikz}

\usepackage[pagebackref,breaklinks,colorlinks]{hyperref}

\usepackage[capitalize]{cleveref}
\crefname{section}{Sec.}{Secs.}
\Crefname{section}{Section}{Sections}
\Crefname{table}{Table}{Tables}
\crefname{table}{Tab.}{Tabs.}

\def\wacvPaperID{1242} %
\def\confName{WACV}
\def\confYear{2024}

\begin{document}

\title{Efficient Semantic Matching with Hypercolumn Correlation}

\author{Seungwook Kim \hspace{0.8cm} Juhong Min \hspace{0.8cm} Minsu Cho \vspace{1.5mm} \\
Pohang University of Science and Technology (POSTECH), South Korea \vspace{1.5mm}\\
\small
\href{http://cvlab.postech.ac.kr/research/HCCNet}{\url{http://cvlab.postech.ac.kr/research/HCCNet}}
}
\maketitle

\begin{abstract}
Recent studies show that leveraging the match-wise relationships within the 4D correlation map yields significant improvements in establishing semantic correspondences - but at the cost of increased computation and latency.
In this work, we focus on the aspect that the performance improvements of recent methods can also largely be attributed to the usage of multi-scale correlation maps, which hold various information ranging from low-level geometric cues to high-level semantic contexts.
To this end, we propose \methodName, an efficient yet effective semantic matching method which exploits the full potential of multi-scale correlation maps, while eschewing the reliance on expensive match-wise relationship mining on the 4D correlation map.
Specifically, \methodName performs feature slicing on the bottleneck features to yield a richer set of intermediate features, which are used to construct a hypercolumn correlation.
\methodName can consequently establish semantic correspondences in an effective manner by reducing the volume of conventional high-dimensional convolution or self-attention operations to efficient point-wise convolutions.
\methodName demonstrates state-of-the-art or competitive performances on the standard benchmarks of semantic matching, while incurring a notably lower latency and computation overhead compared to the existing SoTA methods.

\end{abstract}

\section{Introduction}
\label{sec:intro}

Semantic correspondence is the task of establishing correspondences between two images depicting different instances of the same semantic category.
While visual correspondence itself is a fundamental computer vision task used for 3D reconstruction, visual localization and object recognition~\cite{forsyth:hal-01063327}, semantic correspondence has enabled further diverse applications, including semantic label/edit transfer~\cite{Mu2022coordgan,Peebles2022Gangealing}, unsupervised object discovery/localization~\cite{cho2015unsupervised}, and few-shot classification/segmentation~\cite{Kang_2021_ICCV, min2021hypercorrelation, Kang_2022_CVPR, hong2022cost}.
While the recent success of deep neural networks in keypoint detection~\cite{detone18superpoint, Barroso-Laguna2019ICCV} and feature descriptor extraction~\cite{tian2019sosnet, jerome2019r2d2} have shown significant improvements, 
the task of semantic correspondence remains challenging due to the presence of intra-class variations~\cite{rocco2018neighbourhood, min2020dhpf, min2021chm, jeon2020guided, cho2021semantic, liu2020semantic, swkim2022tfmatcher}.

\begin{figure}[t]
    \begin{center}
    \includegraphics[width=1.0\linewidth]{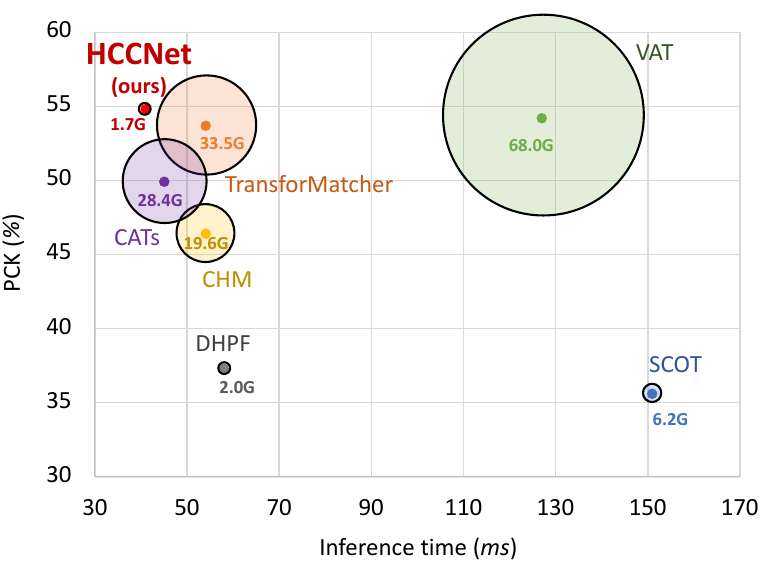}
    \end{center}
     \vspace{-2.5mm}
    \caption{\textbf{PCK performance (y-axis) {\em vs.} inference time (x-axis) on SPair-71k dataset.} The area of each bubble is proportional to FLOPs of a model. We demonstrate that the proposed HCCNet outperforms existing state-of-the-art methods in terms of accuracy, efficiency, and scalability while being much simpler design than previous work~\cite{cho2021semantic, hong2022cost, liu2020semantic, min2021chm, min2020dhpf, swkim2022tfmatcher}.}
    \vspace{-7.0mm}
    \label{fig:teaser}
\end{figure}

Among many effective learning-based methods that have been proposed by building on the efficacy of convolutional neural networks~\cite{jeon2018parn, rocco18weak, jeon2020guided,min2019hyperpixel, min2020dhpf}, a representative branch was largely inspired by the idea of learning geometric matching with high-dimensional convolution~\cite{li2020correspondence,rocco2020sparse, li20dualrc, min2021chm}, where convolutional layers are applied to the correlation map such that the certain unique matches would support the neighboring ambiguous matches.
Noting that the convolution may suffer from inherent limitations of static and local transformations of the correlation map, current state-of-the-art methods propose to leverage self-attention to learn the global match-wise relations~\cite{cho2021semantic, swkim2022tfmatcher, hong2022cost}.

While leveraging the global interactions within a correlation map has shown to be highly effective, we suggest that the superior performance of today's state-of-the-art methods can also be attributed to the usage of multi-scale correlation maps (Section~\ref{sec:experiments}, Table~\ref{tbl:ablate_single}).
This is because semantic correspondences between images having large intra-class variations may occur at different feature levels, from local patterns and geometries (shallow) to invariant semantics and context (deep).
It has also been demonstrated in other areas such as few-shot 
segmentation~\cite{min2021hypercorrelation, hong2022cost} that leveraging multi-layer correlation maps shows improvements over using just a single correlation map.

In this work, we shift our focus away from mining global match-wise relations, to \textit{better} leveraging the multi-scale correlation maps holding various semantics. 
To this end, we introduce an efficient yet effective semantic matching method, \methodName, which carries out \textit{feature slicing} to yield a richer set of equi-channel intermediate features from the backbone network, for constructing amplified multi-scale correlation maps.
These multi-scale correlation maps are concatenated along the channel dimension to obtain a hypercolumn correlation.
We finally perform a fast and efficient point-wise channel aggregation to output a refined correlation map for semantic keypoint transfer.
The results demonstrate that our method surpasses existing state of the arts in terms of accuracy and efficiency despite its simple, straightforward design, as illustrated in Fig.~\ref{fig:teaser}.

The contributions of our work is threefold:

\begin{itemize}
    \item We introduce \methodName, a novel semantic matching learner that leverages various semantics of multi-scale correlations to establish reliable correspondences,
    \item We propose feature slicing, a method to yield rich feature slices from intermediate features to construct an informative hypercolumn correlation,
    \item The rich hypercolumn correlation enables \methodName to reduce the volume of high-dimensional convolution or self-attention operations to point-wise channel aggregation, incurring notably lower computation and latency overhead while exhibiting SoTA or competitive performance on semantic correspondence.
\end{itemize}

\section{Related Work}
\label{sec:related_work}

\begin{figure*}
    \begin{center}
        \includegraphics[width=1.0\linewidth]{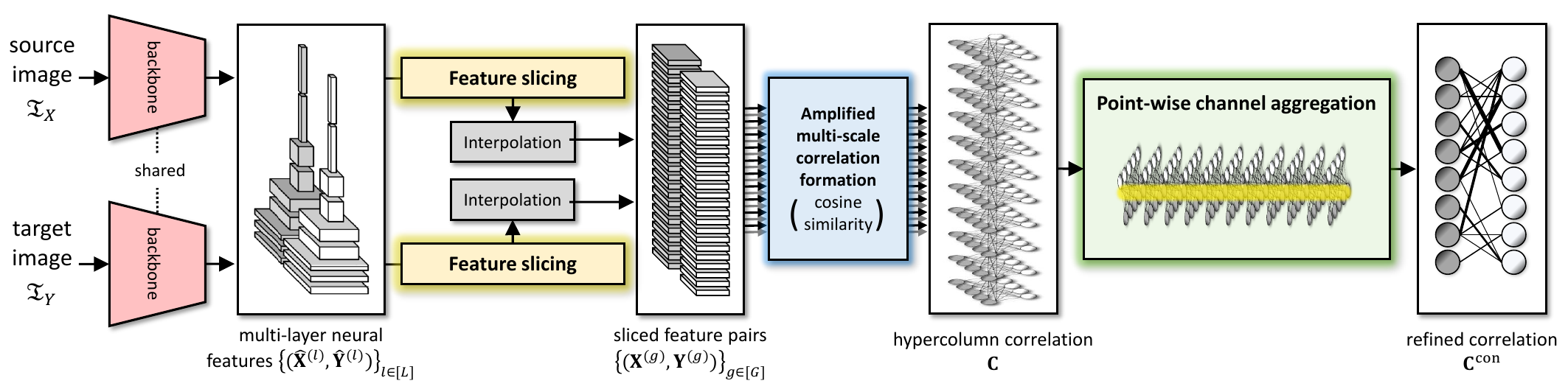}
    \end{center}
    \vspace{-4.0mm}
      \caption{\textbf{Overview of \methodName.}
      The intermediate feature maps extracted from an image pair are first sliced, and are used to compute a consequently amplified multi-channel correlation map.
      We then identify and exploit the position-specific inter-correlation consensuses to provide the refined single-channel correlation map.
      We construct a dense flow field from the refined correlation map, which can be used to transfer given source keypoints to the target image to supervise \methodName using ground-truth keypoint pair annotation.
    }

    \label{fig:overview}
\end{figure*}

\smallbreak
\noindent \textbf{Leveraging multi-layer features and correlations for correspondence.}
For CNNs trained on the task of object recognition, the shallower layers learn geometric cues such as edge or color, and the deeper layers learn semantic cues of the object~\cite{Gatys2015TextureSU, Gatys2016ImageST}. 
This characteristic of hierarchical features of CNNs have been applied to the task of establishing correspondences between images.
Specifically, HPF~\cite{min2019hyperpixel} and its follow-up work, DHPF~\cite{min2020dhpf}, propose to represent images using hyperpixels by leveraging a number of layers selected among early to late layers of the feature extractor.
COLD~\cite{lee2021distill} performs weighted summation on the intermediate feature maps to yield a distilled feature map pair.
More recently, TransforMatcher~\cite{swkim2022tfmatcher} proposed to use multi-layer correlation maps, but only as features of each match position to be processed by match-to-match attention, without explicitly leveraging their consensus.

We focus on the aspect that semantic correspondences between images may occur at different feature levels depending on the image pair.
To this end, we propose feature slicing to yield amplified multi-scale correlation maps to maximize the potential of the constructed hypercolumn correlation, on which we perform point-wise channel aggregation to exploit the various semantics of the correlation maps.
We empirically demonstrate the superiority of our approach over concatenating or summing multi-level features to construct a single correlation map.

\smallbreak
\noindent \textbf{Consensus-based semantic correspondence.}
The task of semantic correspondence aims to establish correspondences between images of the same category but of different instances.
While various CNN-based methods have been introduced to tackle this challenging problem~\cite{min2020dhpf, li2020correspondence, jeon2020guided, min2021chm}, with the recent advent of transformer-based architectures for visual tasks~\cite{dosovitskiy2020vit}, transformer-based methods have demonstrated superior abilities to establish accurate semantic correspondences~\cite{cho2021semantic, swkim2022tfmatcher, hong2022cost}.

Among these approaches, NCNet~\cite{rocco2018neighbourhood} coined the idea of exploiting the local neighborhood consensus within the correlation map using high-dimensional convolutional networks.
The efficacy of this approach motivated follow-up work to better exploit the neighborhood consensus to obtain reliable correspondences~\cite{li2020correspondence,rocco2020sparse, li20dualrc, min2021chm} using high-dimensional CNNs.
However, these methods suffer from the inherent limitations of CNNs \ie, local and static feature transformation.
To alleviate these issues, the current SoTA methods on semantic correspondence exploit the
\textit{dynamic global} match-wise consensus in the correlation map~\cite{cho2021semantic, swkim2022tfmatcher, hong2022cost} by building on self-attention mechanisms.

Albeit its efficacy, the endeavor to mine local or global match-wise relationships in the correlation map incurs high computation overhead.
In this work, we propose to leverage hypercolumn correlation built from multi-scale correlation maps instead.
As the multi-scale correlation maps are derived from feature maps of largely varying receptive fields, \methodName implicitly considers the neighborhood consensus when performing feature matching after the point-wise channel aggregation despite its efficiency.

\smallbreak
\noindent \textbf{Attention for feature aggregation.}
Attention mechanisms enable neural networks to concentrate on the most relevant features, which has shown to be effective across many visual tasks such as object recognition and semantic segmentation~\cite{chen2016attention, hu2018squeeze, wang2017residual, woo2018cbam, yu2018bisenet}. 
SENet~\cite{hu2018squeeze} exploits the channel-wise relationships by introducing the Squeeze-and-Excitation module.
CBAM~\cite{woo2018cbam} combines the spatial and channel attention in a compact block.
Bisenet~\cite{yu2018bisenet} suggests a lightweight module for channel-wise attention.
Noting the effectiveness of employing attention-based mechanisms, more recent work propose to leverage attention to aggregate features.
GFF~\cite{li2020gated} selectively fuses features from multiple levels using a gating mechanism in a fully connected manner.
BPNet~\cite{nie2020bidirectional} uses an add-multiply-add fusion block to first add and multiply features from different levels separately, and then adds these two output features together.
The weighted addition of features proposed in COLD~\cite{lee2021distill} is also a form of attentive feature aggregation.

In this work, instead of fusing features extracted across the feature extractor, we propose to aggregate the channels of the hypercolumn \textit{correlation} in a point-wise manner to obtain a refined correlation map for efficient and effective correspondence establishment.

\section{\methodName for semantic correspondence}
\label{sec:method}

We first provide an overview of how \methodName establishes semantic correspondences.
Given an image pair to match, we yield a set of intermediate feature maps using the backbone feature extractor network.
These intermediate feature maps are bilinearly interpolated to the same spatial size, on which we perform \textit{feature slicing} to yield a larger number of equi-channel feature maps.
Each corresponding feature slice pair is used to calculate a single-channel correlation map, collectively yielding a set of multi-scale correlation maps.
The multi-scale correlation maps are concatenated along the channel dimension to construct a \textit{hypercolumn correlation}, on which we perform efficient point-wise convolution to aggregate the channels to output a refined correlation map.
This refined correlation map is used to construct a dense flow field, which is used to transfer the given keypoints from the source images to the target image to establish correspondences between the image pair.
Figure \ref{fig:overview} illustrates the overall architecture of our method.

\subsection{Feature slicing}

We utilize an ImageNet-pretrained ResNet-101~\cite{he2016deep, deng2009imagenet} as our feature extractor.
To maximize the number of correlation maps and thus the visual cues to consider, we extract features from all bottleneck layers of \texttt{conv3\_x}, \texttt{conv4\_x}, and \texttt{conv5\_x} blocks for a given pair of images $(\mathcal{I}_{X},\mathcal{I}_{Y})$.
The multiple intermediate features are bilinearly interpolated to achieve the same (flattened) spatial dimension of $HW$; this dimension is $\frac{1}{16}$ of the input image resolution, thereby creating a set of features $\{(\hat{\mathbf{X}}^{(l)}, \hat{\mathbf{Y}}^{(l)})\}_{l \in [L]}$ where $\hat{\mathbf{X}}^{(l)}, \hat{\mathbf{Y}}^{(l)} \in \mathbb{R}^{HW \times C^{(l)}}$ represent the feature pair at layer $l$, $[L] \coloneqq \{i\}_{i=1}^{L}$ represents a set of bottleneck layer indices, and $C^{(l)}$ indicates the channel size at layer $l$.

Previous related studies~\cite{cho2021semantic, min2021chm} directly compute cosine similarity on extracted intermediate backbone feature pairs to form correlations, \ie, $\hat{\mathbf{X}}^{(l)} \cdot \hat{\mathbf{Y}}^{(l)\top}$.
However, such an approach could overlook {\em rich channel-wise information} of {\em high-dimensional backbone feature vectors} which potentially helps form richer correlation maps for the model to analyze.
To address this issue, we introduce \textit{feature slicing}, which slices each intermediate feature map, $\hat{\mathbf{X}}^{(l)}$ or $\hat{\mathbf{Y}}^{(l)}$, into multiple slices to provide a larger number of feature pairs by increasing the number of features to match.
Specifically, we view each feature map at every layer as a composition of multiple sub-features concatenated along the channel dimension: $\hat{\mathbf{X}}^{(l)} \coloneqq \text{concat}_{g \in G^{(l)}}\left[ \mathbf{X}^{(g)} \right]$ for all $l \in [L]$, where $G^{(l)}$ is the number of slices used to divide feature map $\mathbf{X}^{(l)}$.
This interpretation provides us with a more diverse set of visual features for the subsequent matching network to establish reliable matches, which we denote as $\{(\mathbf{X}^{(g)}, \mathbf{Y}^{(g)})\}_{g \in [G]}$ where $L < G$.

\subsection{Hypercolumn correlation construction}
To establish dense input pair-wise matches, we first calculate the cosine similarity between every possible position pairs between the feature maps. Specifically, for each group $g \in [G]$, we compute the dense, richer ($L < G$) correlation matrix $\mathbf{C}_{:, :, g} \in \mathbb{R}^{HW \times HW}$ as follows:
\begin{align}
    \mathbf{C}_{\mathbf{x}, \mathbf{y}, g} = \frac{\mathbf{X}^{(g)}_{\mathbf{x},:} \cdot \textbf{Y}^{(g)\top}_{\mathbf{y},:}}{\|\mathbf{X}^{(g)}_{\mathbf{x},:}\|_{2}\| \mathbf{Y}^{(g)}_{\mathbf{y},:} \|_{2}},
\end{align}
where $\mathbf{x}, \mathbf{y} \in \mathbb{R}^{2}$ refer to the 2-dimensional spatial positions of the feature maps corresponding to the image pair of $\mathcal{I}_{X}$ and $\mathcal{I}_{Y}$ respectively.
While some existing methods~\cite{min2021chm, swkim2022tfmatcher} apply ReLU on top of correlation maps for non-negativity, we propose that the negative correlation scores also provide important cues for reliable correspondences, as empirically evidenced by better performance.
After calculating the correlation maps for $G$ feature slice pairs, we stack them along the channel dimension to produce the final hypercolumn correlation, denoted by $\mathbf{C} \in \mathbbm{R}^{HW \times HW \times G}$.
This approach enables us to consider a diverse set of intermediate feature pairs and provides richer information for the subsequent matching network to establish reliable matches.

\subsection{Point-wise channel aggregation}
When convolutional neural networks are trained on the task of object recognition, the feature representations become increasingly explicit about the object information along the processing hierarchy~\cite{Gatys2015TextureSU, Gatys2016ImageST}.
Specifically, low-layer features contain more detailed information such as edge or colour, while higher-layer features contain more semantic information with higher invariance~\cite{DU2020108, zeiler2013visualize}.
Pertaining to the task of semantic matching, it is unsure at which feature layer the correlation is likely to be the most accurate, as the images depict different instances of the same class.
Therefore, depending on the content of the given image pair, it may be beneficial to rely more on the lower-layer features, or rather on the higher-layer features.
Now that we have a hypercolumn correlation $\mathbf{C}$ which holds the correlation information obtained from different layers of the backbone feature extractor, we aim to analyze the channels in a point-wise manner to aggregate the channels in order to yield the final refined correlation matrix.

It is crucial to ensure that a flow field has reliable and consistent information after analyzing different visual cues to establish reliable correspondences.
To facilitate this, we aim to analyze the channels of the hypercolumn correlation for each spatial position $(\mathbf{x}, \mathbf{y})\in \mathbb{R}^{4}$ in $\mathbf{C}$ to focus on or to downweight certain visual cues in aggregating the channels.
Therefore, we collect match scores from different visual aspects of geometric and semantic cues and perform point-wise convolution as follows:
\begin{align}
    \Phi(\mathbf{C}; \mathbf{W}_{\text{hid}})_{\mathbf{x}, \mathbf{y}, :} \coloneqq \mathbf{C}_{\mathbf{x}, \mathbf{y}, :} \mathbf{W}_{\text{hid}} \in \mathbb{R}^{D_{\text{hid}}},
\end{align}
where $\mathbf{W}_{\text{hid}} \in \mathbb{R}^{G \times D_{\text{hid}}}$ is a learnable weight matrix.
To enhance the correlation consensus with better representational power, we process the correlation map $\mathbf{C}$ using two correlation consensus networks together with an intermediate hyperbolic tangent non-linearity $\zeta$:
\begin{align}
    (\mathbf{C}^{\text{con}})_{\mathbf{x}, \mathbf{y}} &\coloneqq \Phi(\zeta(\Phi(\mathbf{C}; \mathbf{W}_{\text{hid})})); \mathbf{W}_{\text{out}})_{\mathbf{x}, \mathbf{y}} \\
    &= \zeta(\mathbf{C}_{\mathbf{x}, \mathbf{y}, :} \mathbf{W}_{\text{hid})}) \mathbf{W}_{\text{out}} \in  \mathbb{R},
\end{align}
where $\mathbf{W}_{\text{out}} \in \mathbb{R}^{D_{\text{hid}} \times 1}$ is learnable matrix that squeezes multiple channels to provide a single, refined correlation map for the subsequent flow field formation.

Note that we employ a simple yet effective approach that leverages hypercolumn correlation, striking a balance between efficacy and efficiency without resorting to overly complex methodologies \eg match-wise relation mining, for visual correspondence.
In Section~\ref{sec:experiments}, we present empirical evidence that highlights the efficacy of our method despite its straightforward nature, showing that it surpasses existing methods without relying on computationally-demanding techniques, \eg, high-dimensional convolutions~\cite{min2021chm, min2021cpchm}, cost aggregations~\cite{cho2021semantic}, or Hough matching~\cite{min2019hyperpixel, min2020dhpf}.
By avoiding such complicated methodologies and instead relying on straightforward, simple design by leveraging pretrained backbone features, we pave the way for more accessible, scalable, and practical solutions to the problem of visual correspondence.

\subsection{Flow field formation and keypoint transfer}
For fine-grained flow field formation, the aggregated correlation matrix $\mathbf{C}^{\text{con}}$ is then upsampled via a 4-dimensional upsampling function that provides $\mathbf{C}^{\text{out}} \in \mathbb{R}^{\bar{H}\bar{W} \times \bar{H}\bar{W}}$ where $\bar{H} = 4H$ and $\bar{W} = 4W$, which corresponds to $\frac{1}{4}$ the size of the original image.
We use the output correlation tensor $\mathbf{C}^{\text{out}}$ to form a dense flow field between the source and target image for keypoint transfer.
First, we normalize the the output correlation map by applying kernel soft-argmax~\cite{lee2019sfnet} as follows:
\begin{align}
    \mathbf{C}^{\text{norm}}_{\mathbf{x}, \mathbf{y}} = \frac{\text{exp}(\mathbf{G}_{\mathbf{y}}^\mathbf{p}\mathbf{C}^{\text{out}}_{\mathbf{x}, \mathbf{y}})}{\sum_{\mathbf{m} \in [\bar{H}] \times [\bar{W}]}\text{exp}(\mathbf{G}_{\mathbf{m}}^\mathbf{p}\mathbf{C}^{\text{out}}_{\mathbf{x}, \mathbf{m}})} \in \mathbb{R},
\end{align}
where $\mathbf{G}^\mathbf{p} \in \mathbb{R}^{\bar{H} \times \bar{W}}$ is a 2D Gaussian kernel centered on $\mathbf{p} = \text{arg max}_{\mathbf{y}} \mathbf{C}^{\text{out}}_{\mathbf{x}, \mathbf{y}}$, to suppress noisy correlation values in the correlation map.
The normalized correlation tensor $\mathbf{C}^{\text{norm}}$ encodes a set of probability simplexes from each source feature position to the target feature positions.
We then transfer all the coordinates on the dense regular grid $\mathbf{P}_{X} \in \mathbb{R}^{\bar{H}\bar{W} \times 2}$ of source image $\mathcal{I}_{X}$ to obtain their corresponding coordinates $\mathbf{\hat{P}}_{Y} \in \mathbb{R}^{\bar{H}\bar{W} \times 2}$ on target image $\mathcal{I}_{Y}$:
\begin{align}
    (\mathbf{\hat{P}}_{Y})_{\mathbf{x},:} = \sum_{(\mathbf{y})\in [\bar{H}] \times [\bar{W}]} \mathbf{C}^{\text{norm}}_{\mathbf{x},\mathbf{y}}(\mathbf{P}_{X})_{\mathbf{y},:} \in \mathbb{R}^{2},
\end{align}
forming a dense flow field.
Using this dense flow field, we can perform keypoint transfer as follows.
Given a keypoint $\mathbf{k}^{X} = (x_k,y_k)$, we define a soft sampler $\textbf{W}^{(k)} \in \mathbb{R}^{\bar{H} \times \bar{W}}$:
\begin{align}
\mathbf{W}_{ij}^{(k)} = \frac{\text{max}(0, \tau - \sqrt{(x_k - j)^2 + (y_k - i)^2})}{\sum_{i'j'} \text{max}(0, \tau - \sqrt{(x_k - j')^2 + (y_k - i')^2})},
\end{align}
where $\tau$ is a distance threshold, and $\sum_{ij} \mathbf{W}_{ij}^{(k)} = 1$.
The above equation shows that the soft sampler samples each transferred keypoint $(\mathbf{\hat{P}}_{Y})_{ij}$ by assigning weights which are inversely proportional to the distance to $\mathbf{k}^{X}$.
Using this soft sampler, we assign a match to the keypoint $\mathbf{k}^{X}$ as $\mathbf{k}^{Y} = \sum_{(i,j) \in [\bar{H}] \times [\bar{W}]} (\mathbf{\hat{P}}_{Y})_{ij:}\mathbf{W}_{ij}^{(k)}$, being able to establish sub-pixel-wise accurate correspondences.
\begin{table*}
\centering
\scalebox{0.95}{
\begin{tabular*}{\textwidth}{l@{\extracolsep{\fill}}ccccccccc}
                \toprule
                \multirow{3}{*}{Method} & \multicolumn{2}{c}{SPair-71k} & \multicolumn{2}{c}{PF-PASCAL} & \multicolumn{2}{c}{PF-WILLOW} & \multirow{3}{*}{\shortstack{time\\(\emph{ms})}} & \multirow{3}{*}{\shortstack{memory\\(GB)}} &\multirow{3}{*}{\shortstack{FLOPs\\(G)}}\\
                
                & \multicolumn{2}{c}{@$\alpha_{\text{bbox}}$} & \multicolumn{2}{c}{@$\alpha_{\text{img}}$} & @$\alpha_{\text{bbox-kp}}$ & @$\alpha_{\text{bbox}}$ \\ 
                 
                 & 0.1 (F) & 0.1 (T) & 0.05 (F) & 0.1 (F) & 0.1 (T) & 0.1 (T)\\
                 
                 \midrule

                 HPF~\cite{min2019hyperpixel}      &  28.2 & - & 60.1 & 84.8 & 74.4 & - & 63 & - & -\\
                 
                 SCOT~\cite{liu2020semantic} & 35.6 & - & 63.1 & 85.4 & \textbf{76.0} & - & 151 & 4.6 & \underline{6.2} \\

                 DHPF~\cite{min2020dhpf}           & {37.3} & 27.4  & {75.7} & {90.7} & 71.0 & 77.6 & 58 & 1.6 & \underline{2.0}\\
                 
                 DHPF$\dagger$ ~\cite{min2020dhpf}         & 39.4 & -  & - & - & - & - & 58 & 1.6 & \underline{2.0}\\
                 
                 NC-Net$^\textrm{{*}}$~\cite{rocco2018neighbourhood} & - & - & - & 81.9 & - & -& 222 & \underline{1.2} & 44.9\\
                 
                 DCC-Net$^\textrm{{*}}$~\cite{huang2019dynamic} &  - & - & - & 83.7 & - & - & 567 & 2.7 & 47.1\\
                 
                 ANC-Net~\cite{li2020correspondence} &  - & 28.7 & - & 86.1 & - & - & 216 & \textbf{0.9} & 44.9\\
                 
                 PMD~\cite{li2021pmdnet} &  37.4 & - & - & 90.7 & \underline{75.6} & - & - & - & -\\
                 
                 CHMNet~\cite{min2021chm}   & {46.4} & {30.1} & 80.1 & {91.6} & 69.6 & \underline{79.4} & 54 & 1.6 & 19.6\\
                 
                 PMNC~\cite{lee2020pmnc} &  {50.4} & - & \textbf{82.4} & 90.6 & - & - & - & - & -\\
                 
                 MMNet~\cite{zhao2021multi} &  40.9 & - & 77.6 & 89.1 & - & - & 86 & - & -\\
                 
                 CATs~\cite{cho2021semantic}   & 43.5 & - & - & - & - & - & \underline{45} & 1.6 & 28.4\\
                 
                 CATs$\dagger$~\cite{cho2021semantic}   & 49.9 & 27.1 & 75.4 & \textbf{92.6} & 69.0 & 79.2 & \underline{45} & 1.6 & 28.4\\
                
                PWarpC-NC-Net~\cite{Truong2022ProbabilisticWC}   & 52.0 & \textbf{37.1} & 67.8 & 82.3 & - & 76.2 & - & - & -  \\
                
                 TransforMatcher~\cite{swkim2022tfmatcher}   & 50.2 & \underline{30.5} & 78.9 & 90.5 & 66.7 & 75.1 & 54 & 1.6 & 33.5 \\
                 
                 TransforMatcher$\dagger$~\cite{swkim2022tfmatcher}   & 53.7 & {30.1} & \underline{80.8} & {91.8} & 65.3 & 76.0 & 54 & 1.6 & 33.5  \\
                 
                 VAT$\dagger$~\cite{hong2022cost}   & \underline{54.2} & - & 78.2 & 92.3 & - & \textbf{81.6} & 127 & 3.6 & 68.0  \\

                 \midrule
                 \midrule
                 
                 \methodName (ours) & 53.9 & 29.6 & 80.2 & \underline{92.4} & 65.3 & 74.5 & \textbf{30} & 2.0 & \textbf{1.7} \\
                 \methodName $\dagger$ (ours) & \textbf{54.8} & 29.7 & 80.2 & \underline{92.4} & 65.5 & 74.5 & \textbf{30} & 2.0 & \textbf{1.7}  \\
                 
                 \bottomrule
        \end{tabular*}
        }
        \vspace{2.0mm}
        \caption{\textbf{Performance on standard benchmarks of semantic matching.} All the methods reported in the above table uses a pretrained ResNet-101 model as the feature extractor.
        The first group of methods were trained with weak supervision (image pair annotations), and the second group of methods were trained using strong supervision (keypoint pair annotations).
        Models with $\textrm{{*}}$ are retrained using keypoint annotations from ANC-Net~\cite{li2020correspondence}.
        $\dagger$ indicates the use of data augmentation during training.
        Numbers in bold indicate the best performance, followed by the underlined numbers.
        }
        \label{tbl:main_table}
        \vspace{1.0mm}
\end{table*}

\subsection{Training objective}

For each training image pair with ground-truth correspondences $\mathcal{M} = \{(\hat{\mathbf{k}}^{X}_m, \hat{\mathbf{k}}^{Y}_m)\}_{m=1}^M$, we apply the aforementioned keypoint transfer method on the given source keypoints to obtain predicted target keypoints.
This results in a set of predicted correspondences $\{(\hat{\textbf{k}}^{X}_m, \textbf{k}^{Y}_m)\}_{m=1}^M$ by assigning a match $\textbf{k}^{Y}_m$ to each keypoint $\hat{\textbf{k}}^{X}_m$ in the source image.
We formulate our training objective to minimize the average Euclidean distance between the predicted target keypoints and the ground-truth target keypoints as follows:
\begin{align}
\mathcal{L} = \frac{1}{M}\sum_{m=1}^M \|\mathbf{k}^{Y}_m - \hat{\mathbf{k}}^{Y}_m \|_{2}^{2}.
\end{align}

\section{Experiments}
\label{sec:experiments}

\begin{table*}[!htb]
\centering
\scalebox{0.7}{
\begin{tabular}{lccccccccccccccccccc}
                \toprule
                Methods & aero & bike & bird & boat & bottle & bus & car & cat & chair & cow & dog & horse & mbike & person & plant & sheep & train & tv & all \\

                 \midrule
                 
                NC-Net~\cite{rocco2018neighbourhood} & 23.4 & 16.7 & 40.2 & 14.3 & 36.4 & 27.7 & 26.0 & 32.7 & 12.7 & 27.4 & 22.8 & 13.7 & 20.9 & 21.0 & 17.5 & 10.2 & 30.8 & 34.1 & 20.6 \\

                HPF~\cite{min2019hyperpixel} & 25.2 & 18.9 & 52.1 & 15.7 & 38.0 & 22.8 & 19.1 & 52.9 & 17.9 & 33.0 & 32.8 & 20.6 & 24.4 & 27.9 & 21.1 & 14.9 & 31.5 & 35.6&  28.2\\
                 
                SCOT~\cite{liu2020semantic} & 34.9 & 20.7 & 63.8 & 21.1 & 43.5 & 27.3 & 21.3 & 63.1 & 20.0 & 42.9 & 42.5 & 31.1 & 29.8 & 35.0 & 27.7 & 24.4 & 48.4 & 40.8 & 35.6\\
                 
                DHPF~\cite{min2020dhpf} & 38.4 & 23.8 & 68.3 & 18.9 & 42.6 & 27.9 & 20.1 & 61.6 & 22.0 & 46.9 & 46.1 & 33.5&  27.6&  40.1&  27.6&  28.1&  49.5&  46.5&  37.3\\
                 
                CHMNet~\cite{min2021chm}   & 49.6 & 29.3 & 68.7 & 29.7 & 45.3 & 48.4 & 39.5 & 64.9 & 20.3 & 60.5 & 56.1 & 46.0 &  33.8&  44.2&  38.9&  31.3&  72.2&  55.6&  46.4 \\
                 
                PMNC~\cite{lee2020pmnc} & 54.1 & {35.9} & \textbf{{74.9}} & 36.5 & 42.1 & 48.8 & 40.0 & \textbf{72.6} & 21.1 & 67.6 & {58.1} & 50.5 & 40.1 & \textbf{{54.1}} & 43.3 & {35.7} & {74.5}  & 59.9 & {50.4}\\
                 
                MMNet~\cite{zhao2021multi} & 43.5 & 27.0 & 62.4 & 27.3 & 40.1 & 50.1 & 37.5 & 60.0 & 21.0 & 56.3 & 50.3 & 41.3 & 30.9 & 19.2 & 30.1 & 33.2 & 64.2 & 43.6 & 40.9\\
                
                CATs~\cite{cho2021semantic}   & 46.5 & 26.9 & 69.1 & 24.3 & 44.3 & 38.5 & 30.2 & 65.7 & 15.9 & 53.7 & 52.2 & 46.7 & 32.7 & 35.2 & 32.2 & 31.2 & 68.0 & 49.1 & 43.5\\
                 
                CATs$\dagger$~\cite{cho2021semantic}   & 52.0 & 34.7 & 72.2 & 34.3 & {49.9} & {57.5} & 43.6 & 66.5 & 24.4 & 63.2 & 56.5 & {52.0} & {42.6} & 41.7 & 43.0 & 33.6 & 72.6 & 58.0 & 49.9\\

                 TransforMatcher~\cite{swkim2022tfmatcher} & {54.5} & 33.9 & 72.2 & {38.5} & 47.7 & 55.3 & {45.6} & 65.7 & 25.2 & 62.6 & {58.0} & 47.0 & 40.7 & {44.2} & 43.1 & 35.3 & 71.9 & {61.6} & 50.2 \\
                 TransforMatcher~\cite{swkim2022tfmatcher}$\dagger$  & \underline{59.2} & \underline{39.3} & \underline{73.0} & \underline{41.2} & \underline{52.5} & \textbf{66.3} & 55.4 & {67.1} & \underline{26.1} & {67.1} & 56.6 & 53.2 & {45.0} & 39.9 & 42.1 & 35.3 & 75.2 & \underline{{68.6}} & 53.7 \\                 
                 VAT$\dagger$~\cite{hong2022cost} & 56.5 & 37.8 & \underline{73.0} & 38.7 & 50.9 & 58.2 & 40.9 & \underline{70.5} & 20.3 & \textbf{72.1} & \textbf{61.1} & \underline{57.7} & 45.6 & \underline{48.2} & \textbf{52.4} & \textbf{40.0} & 77.7 & \textbf{71.4} & \underline{54.2} \\
                 \midrule
                 \midrule

                 \methodName & \textbf{59.9} & 39.1 & 71.0 & \textbf{42.1} & 51.6 & 63.4 & \textbf{57.0} & 63.0 & \textbf{26.8} & 63.8 & 59.4 & 54.7 & \textbf{49.4} & 41.0 & 43.0 & 37.6 & \textbf{83.1} & 64.8 & 53.9 \\
                 \methodName $\dagger$  & \textbf{59.9} & \textbf{40.6} & 70.5 & 39.8 & \textbf{55.9} & \underline{65.1} & \underline{56.8} & 66.6 & 25.6 & \underline{69.2} & \underline{59.6} & \textbf{58.7} & \underline{46.7} & 40.3 & \underline{43.6} & \underline{39.6} & \underline{82.2} & 65.4 & \textbf{54.8} \\
                 \bottomrule
        \end{tabular}
        }
        \vspace{1.5mm}
        \caption{\textbf{Classwise PCK on SPair-71k.} All the methods reported in the above table uses a pretrained
        ResNet-101 model as the feature extractor. $\dagger$ indicates the use of data augmentation during training.
        We take results from the methods whose classwise PCK results were provided.
        Numbers in bold indicate the best performance, followed by the underlined numbers.
        }
        \label{tbl:classwise}
        \vspace{-3.5mm}
\end{table*}
\begin{table}[!t]
    \centering
      \begin{tabular*}{0.48\textwidth}{l@{\extracolsep{\fill}}cc}
                \toprule
                \multirow{2}{*}{Method} &  \multicolumn{2}{c}{SPair-71k ($\alpha_{\text{img}}$)} \\
                
                & 0.05 & 0.1\\
                
                \midrule
                
                CHMNet$_\textrm{conv3\_x}$~\cite{min2021chm} & - & 47.0 \\
                CHMNet$_\textrm{multi}$~\cite{min2021cpchm} & - & 51.3 \\

                \midrule
                
                *CATs$_\textrm{conv3\_x}$~\cite{cho2021semantic} & 26.2 & 48.3 \\
                CATs$_\textrm{multi}$~\cite{cho2021semantic} & 27.7 & 49.9 \\

                \midrule
                
                TransforMatcher$_\textrm{concat}$~\cite{swkim2022tfmatcher} & 20.9 & 41.7 \\
                TransforMatcher$_\textrm{mean}$~\cite{swkim2022tfmatcher} & 24.1 & 45.1 \\
                TransforMatcher$_\textrm{multi}$~\cite{swkim2022tfmatcher} & 32.4 & 53.7 \\

                \bottomrule
        \end{tabular*}
        \vspace{1.0mm}
        \caption{\textbf{PCK performance of existing methods when using a single correlation map v.s. multiple correlation maps on the SPair-71k dataset.} 
        The results are taken from their reported results, except for *CATs$_\textrm{conv3\_x}$ which was implemented by us.
        }
        \label{tbl:ablate_single}
        \vspace{-5mm}
\end{table}
\begin{table}[!t]
    \centering
       \begin{tabular*}{0.48\textwidth}{cccccccc}
                \toprule
                \multicolumn{4}{c}{\texttt{conv} used} &  \multicolumn{2}{c}{PF-PASCAL} & \multirow{3}{*}{\shortstack{mem.\\(GB)}} &\multirow{3}{*}{\shortstack{FLOPs\\(G)}}\\
                
                \multirow{2}{*}{2\_x} & \multirow{2}{*}{3\_x}  & \multirow{2}{*}{4\_x}& \multirow{2}{*}{5\_x} & \multicolumn{2}{c}{@$\alpha_{\text{img}}$} & & \\ 
                 
                & & & & 0.05 & 0.1 & & \\
                
                \midrule

                $\times$ & $\times$ & $\times$ & \checkmark & 72.1 & 90.1 & 1.9 & 0.9  \\ 
                
                $\times$ & $\times$ & \checkmark & $\times$ & 79.5 & 91.6 & 1.9 & 1.3 \\

                $\times$ & $\times$ & \checkmark & \checkmark & 80.2 & 91.8 & 1.9 & 1.6 \\

                $\times$ & \checkmark & \checkmark & \checkmark & 80.2 & 92.4 & 2.0 & 1.7 \\

                \checkmark & \checkmark & \checkmark & \checkmark & 79.9 & 92.1 & 2.0 & 1.7 \\
                \bottomrule
        \end{tabular*}
        \vspace{1.5mm}
        \caption{\textbf{Ablation study on the backbone bottleneck features used.}
        The results show that our current setting of using \texttt{conv3\_x} to \texttt{conv5\_x} yields the best results.
        }
        \vspace{-2mm}
        \label{tbl:main_ablation_conv}
\end{table}

\begin{table}[!t]
    \centering
       \begin{tabular*}{0.48\textwidth}{c@{\extracolsep{\fill}}ccccc}
                \toprule
                 \multirow{3}{*}{Slice size}&  \multicolumn{2}{c}{PF-PASCAL} & \multirow{3}{*}{\shortstack{time\\(\emph{ms})}} & \multirow{3}{*}{\shortstack{mem.\\(GB)}} &\multirow{3}{*}{\shortstack{FLOPs\\(G)}}\\
                
                & \multicolumn{2}{c}{@$\alpha_{\text{img}}$} & & & \\ 
                 
                 & 0.05 & 0.1 & & \\
                
                \midrule

                - & 77.3 & 92.2 & 20 & 2.0 & 0.9 \\ 
                512 & 77.0 & 91.9 & 24 & 2.0 & 1.1 \\
                256 & 80.2 & 92.4 & 30 & 2.0 & 1.7 \\
                128 & 80.2 & 92.2 & 43 & 2.2 & 4.0 \\
                64 & 79.5 & 92.5 & 70 & 2.2 & 13.3 \\
                32 & 80.4 & 92.4 & 127 & 2.7 & 50.7 \\
                16 & 79.9 & 91.5 & 290 & 3.8 & 200 \\
                8 & 65.0 & 82.5 & 580 & 5.0 & 798 \\

                \bottomrule
        \end{tabular*}
        \vspace{2.0mm}
        \caption{\textbf{Ablation study on the slice size used.}
        The results show that our current setting of using the chunk size of 256 yields the best trade-off between performance and efficiency. 
        }
        \label{tbl:main_ablation_chunk}
        \vspace{-2.0mm}
\end{table}

\begin{table}[!t]
    \centering
       \begin{tabular*}{0.48\textwidth}{c@{\extracolsep{\fill}}cc}
                \toprule
                 \multirow{3}{*}{Activation function}&  \multicolumn{2}{c}{PF-PASCAL} \\ 
                
                & \multicolumn{2}{c}{@$\alpha_{\text{img}}$}\\ 
                 
                 & 0.05 & 0.1 \\
                
                \midrule

                ReLU & 79.5 & 91.9 \\ 
                Sigmoid & 79.6 & 91.8 \\
                Tanh & \textbf{80.2} & \textbf{92.4} \\
            
                \bottomrule
        \end{tabular*}
        \vspace{2.0mm}
        \caption{\textbf{Ablation study on the non-linear activation function used.}
        Using the Tanh activation function yields the best results, over ReLU or Sigmoid activation functions.
        }
        \label{tbl:main_ablation_nonlinearity}
        \vspace{-4.0mm}
\end{table}

In this section, we evaluate \methodName~against the state-of-the-art methods on the task of semantic matching and discuss the results with in-depth analysis.

\begin{figure*}
    \begin{center}
        \includegraphics[width=0.8\linewidth]{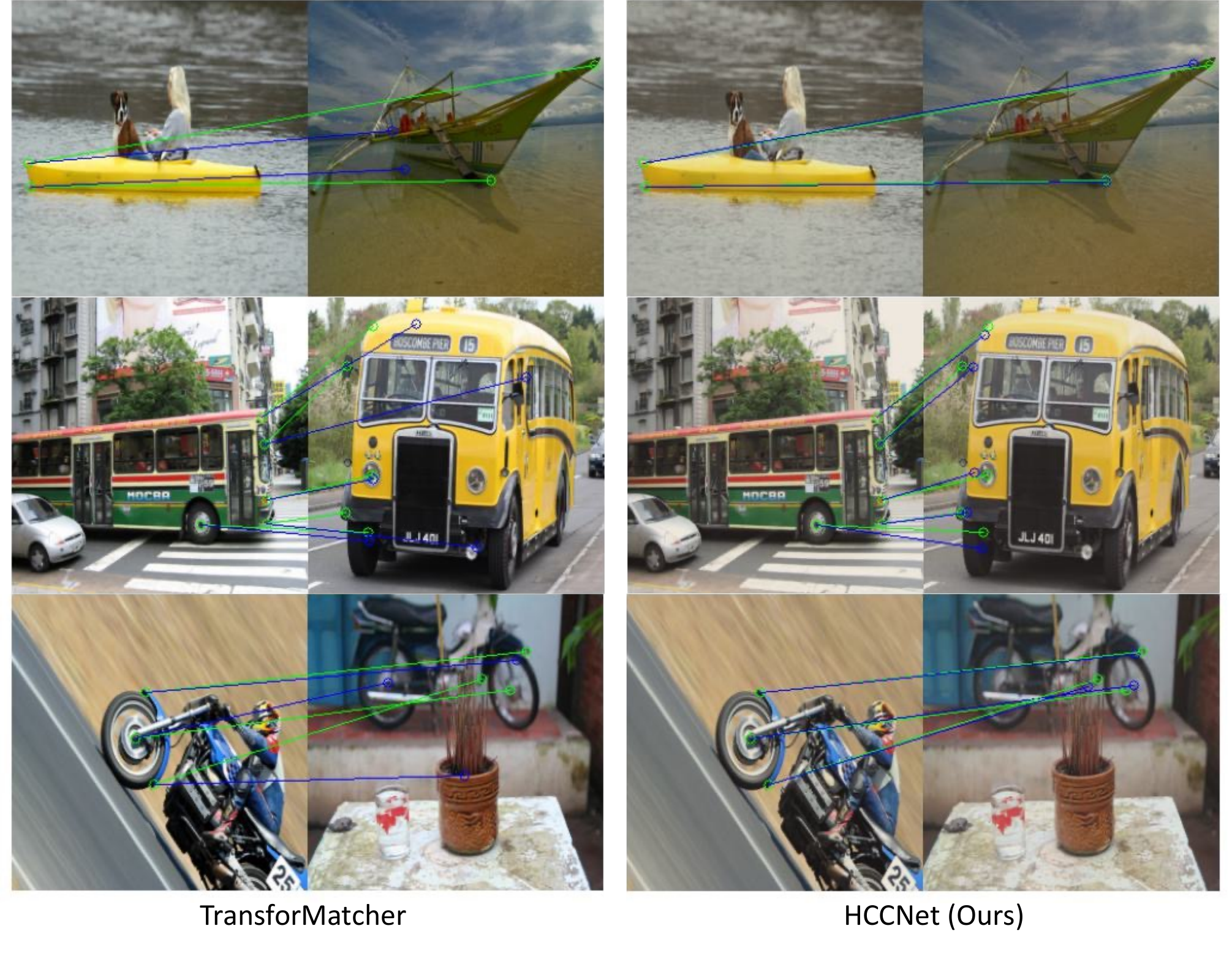}
    \end{center}
    \vspace{-6mm}
      \caption{\textbf{Qualitative comparison of \methodName against TransforMatcher}~\cite{swkim2022tfmatcher}. Green lines represent ground truth correspondences, and blue lines represent predicted correspondences. Best viewed on electronics.
}
\vspace{-3.5mm}
\label{fig:corrcon_qual}
\end{figure*}

\subsection{Evaluation settings}
\smallbreak
\noindent
\textbf{Implementation details.} We use the ImageNet~\cite{deng2009imagenet}-pretrained ResNet-101 model~\cite{he2016deep} as our feature extractor.
The \texttt{conv3\_x}, \texttt{conv4\_x} and \texttt{conv5\_x} layers have 4, 23 and 3 bottleneck layers, respectively; we utilize all these bottleneck layers, and use feature slices with channel dimension of 256 to finally yield $G = 124$ feature slice pairs to construct a 124-layer correlation map for an input image pair ($G=30$). 
We use an image size of $240 \times 240$ for both training and inference, where the feature map dimensions used for correlation computation is $H = W = 15$, and the upsampled $\mathbf{C}^{\text{out}}$ has spatial dimensions of $\bar{H} = \bar{W} = 60$.
Both the number of groups and channel size of linear layers in correlation consensus network are set to 124, \ie, $G = D_{\text{hid}} = 124$. 
\methodName is implemented using PyTorch~\cite{pytorch}, and our network is optimized using the AdamW~\cite{Loshchilov2019DecoupledWD} optimizer with a learning rate of 1e-3 for the correlation network, and 1e-5 for the ResNet-101 feature extractor.

\smallbreak
\noindent
\textbf{Datasets.} 
We evaluate our method on the standard benchmark datasets of semantic matching: PF-PASCAL, PF-WILLOW~\cite{everingham2015pascal} and SPair-71k~\cite{min2019spair} with keypoint-annotated image pairs. 
PF-PASCAL consists of image pairs from the PASCAL VOC 2007 dataset, having the same viewpoint and small scale variations.
PF-PASCAL contains 2,940 / 308 / 299 image pairs for training, validation and testing, respectively.
PF-WILLOW is comprised of four categories of the PASCAL VOC 2007 and Caltech-256 datasets, having center-aligned image pairs with the same viewpoint and small scale variations.
PF-WILLOW contains 900 image pairs for testing only.
SPair-71k consists of image pairs from PASCAL3D+, and PASCAL VOC 2012 datasets, with diverse variations in viewpoint and scale. 
SPair-71k has 53,340 / 5,384 / 12,234 image pairs for training, validation, and testing, respectively.
The results on SPair-71k are much less saturated in comparison to other benchmarks due to its large scale and challenging variations.

\smallbreak
\noindent
\textbf{Evaluation metric.} 
We use the percentage of correct keypoints (PCK) as the evaluation metric.
Given a pair of ground-truth keypoints and our predicted target keypoints, the PCK can be computed as follows:
\begin{align}
\text{PCK}(\mathcal{K}) = \frac{1}{M}\sum_{m=1}^M\mathbbm{1}[\|\mathbf{k}^{Y}_m - \hat{\mathbf{k}}^{Y}_m\| \leq \alpha_{\tau}\cdot\text{max}(w_{\tau},h_{\tau})],
\end{align}
where $w_{\tau}$ and $h_{\tau}$ denotes the width and height thresholds, which are the width and height of either the entire image or the object bounding box, \ie, $\tau \in \{\text{img, bbox-kp, bbox}\}$, and $\alpha_{\tau}$ is a tolerance factor.

\subsection{Quantitative results on semantic matching}

Table \ref{tbl:main_table} illustrates the quantitative results of \methodName in comparison to existing methods on the standard benchmarks of semantic matching. 
To directly demonstrate the efficacy of our method, we report finetuned (F) results, which are trained on the train set of the corresponding dataset. 
To evaluate the cross-dataset generalizability, we report transferred (T) results, where we use a model trained on the train set of PF-PASCAL for evaluation.
It can be seen that \methodName sets a new state of the art on the finetuned (F) setting of the SPair-71k dataset, which is the most challenging semantic matching benchmark, while being competitive on the finetuned (F) setting of the PF-PASCAL dataset, just 0.2\%p below CATs$\dagger$~\cite{cho2021semantic}.
It is noteworthy that \methodName achieves this while incurring notably lower latency and FLOPs compared to existing methods.
On the contrary, \methodName yields subpar outcomes on the transferred (T) settings, which we conjecture is due to \methodName's brittleness to the domain gap between datasets, resulting in inconsistent point-wise channel aggregation.
The classwise PCK results on SPair-71k is shown in Table \ref{tbl:classwise}, and Figure \ref{fig:corrcon_qual} visualizes example qualitative results on the test set of SPair-71K in comparison to TransforMatcher~\cite{swkim2022tfmatcher}.

\subsection{Ablation study and analysis}

\smallbreak
\noindent
\textbf{Effect of using multiple correlation maps in existing methods.} 
Table~\ref{tbl:ablate_single} illustrates the performance of existing methods when using a single correlation map in comparison to using multiple correlation maps.
It is visible that the significant gain in performance is consistent across different methods, substantiating our claim that the efficacy of today's SoTA methods can be largely attributed to the usage of muliple correlation maps\footnote{While TransforMatcher$_\textrm{mean}$ or TransforMatcher$_\textrm{concat}$ use multi-level features, they yield a single correlation map as a result of mean or concatenation of the multi-level features to yield a single feature map pair.
}.

\smallbreak
\noindent
\textbf{Ablation study on the backbone convolutional blocks used.} 
We compare the results of \methodName when extracting bottleneck features from varying convolutional blocks of the backbone network.
The results in Table~\ref{tbl:main_ablation_conv} shows that our current setting of using \texttt{conv3\_x} to \texttt{conv5\_x} strikes the best balance between performance and efficiency.

\smallbreak
\noindent
\textbf{Analysis on the feature slice size.} 
We compare the results of \methodName when using varying sizes of feature slices, or when directly using the bottleneck features for correlation computation.
The results in Table~\ref{tbl:main_ablation_chunk} show that our current setting of using a slice size of 256 yields a favorable balance between performance and efficiency, and the latency and FLOPs increases dramatically with decreasing slice size.

\smallbreak
\noindent
\textbf{Analysis on the non-linear activation function used.} 
Table~\ref{tbl:main_ablation_nonlinearity} shows that using the hyperbolic tangent (Tanh) non-linear activation function yields favorable results in comparison to ReLU or Sigmoid functions.
We conjecture this is because unlike ReLU or Sigmoid, Tanh is capable of representing unlikely matches using negative correlation scores.

\begin{figure}[t]
    \begin{center}
    \includegraphics[width=1.0\linewidth]{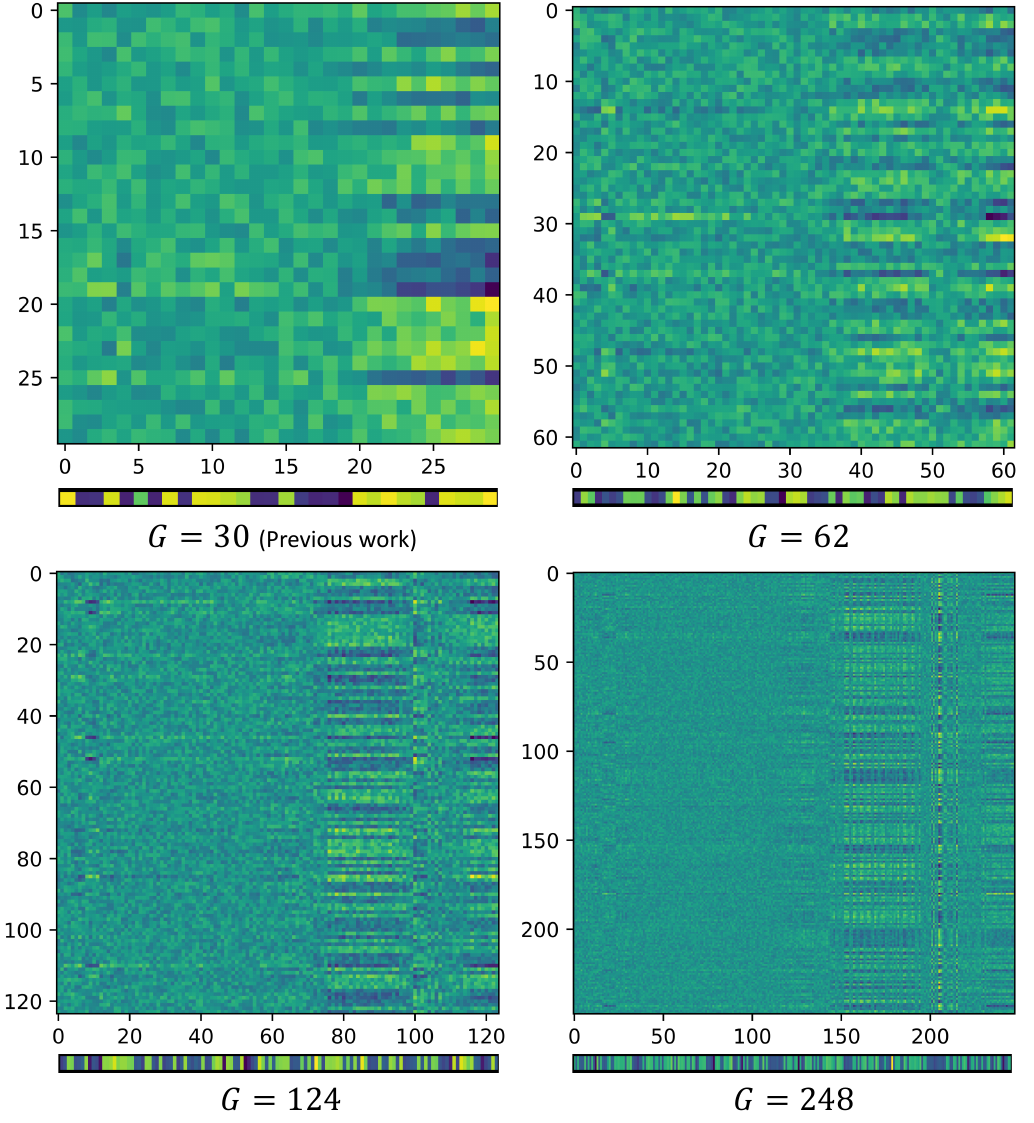}
    \end{center}
     \vspace{-5.0mm}
    \caption{Visualization of learned weight matrices of $\mathbf{W}_{\text{hid}} \in \mathbb{R}^{G \times D_{\text{hid}}}$ (top) and $\mathbf{W}_{\text{out}} \in \mathbb{R}^{D_{\text{hid}} \times 1}$ (bottom) under varying $G = D_{\text{hid}} \in \{30, 62, 124, 248\}$.}
    \vspace{-5.0mm}
    \label{fig:feature_densification}
\end{figure}

\smallbreak
\noindent
\textbf{Feature slicing analysis.} 
To investigate the impact of channel aggregation on the hypercolumn correlation, we visualize learned weight matrices $\mathbf{W}_{\text{hid}}$ and $\mathbf{W}_{\text{out}}$ with four different groups denoted by $G \in \{30, 62, 124, 248\}$\footnote{Note that using $G = 30$ means that feature slicing is not performed, as the total number of intermediate features extracted across the bottleneck layers is already 30.} in Fig.~\ref{fig:feature_densification}.
We observe that the weight magnitudes are notably higher (in yellow) at deeper layers, particularly at \texttt{conv4\_x} and \texttt{conv5\_x}, as opposed to shallower layers.
As we increase the number of groups utilized for feature slicing, we find that the network carries out {\em fine-grained channel selection}, as evidenced by the weight visualization of $\mathbf{W}_{\text{hid}}$, verifying the efficacy of performing position-wise channel aggregation on hypercolumn correlation using diverse visual cues.
Compared to the weight magnitudes of $\mathbf{W}_{\text{hid}}$ that are focused on specific groups, those of $\mathbf{W}_{\text{out}}$ are relatively evenly dispersed in order to effectively aggregate the information from the first channel aggregation to provide a reliable refined correlation map.

We guide the readers to the supplementary for more analyses and experiments of \methodName.

\section{Conclusion}
\label{sec:conclusion}

In this work, we introduced \methodName, an efficient yet effective method to establish semantic correspondences between images. 
Noting that the current trend of mining inter-match relations within the correlation map is computationally demanding, we shifted our focus to \textit{better} leveraging the multi-level correlation maps computed from feature maps of varying receptive fields and visual cues.
Our technical edge lies in the synergistic integration of our proposed feature slicing and point-wise convolution; 
by leveraging feature slicing to yield a richer set of intermediate features, \methodName can effectively establish semantic correspondences while reducing the volume of conventional high-dimensional convolution operations to point-wise convolutions.
Attributing to the eschewal of match-wise relation mining on the correlation map, \methodName incurs notably lower latency and computation overhead while achieving state-of-the-art or competitive performance on the standard benchmarks of semantic correspondence.

\noindent
\paragraph{Acknowledgements.}
This work was supported by the NRF grant (NRF-2021R1A2C3012728 ($50\%$)) and the IITP grants (2022-0-00290: Visual Intelligence for Space-Time Understanding and Generation based on Multi-layered Visual Common Sense ($40\%$), 2019-0-01906:
AI Graduate School Program at POSTECH ($10\%$)) funded by Ministry of Science and ICT, Korea.

\clearpage

{\small
\bibliographystyle{ieee_fullname}
\bibliography{egbib}
}

\setcounter{section}{0}
\setcounter{table}{0}
\setcounter{figure}{0}
\renewcommand{\thesection}{\Alph{section}}
\renewcommand\thefigure{A\arabic{figure}}
\renewcommand{\thetable}{A\arabic{table}}
\clearpage
\title{Efficient Semantic Matching with Hypercolumn Correlation \\
{\it ----- Supplementary Material -----}}

\author{Seungwook Kim \hspace{0.8cm} Juhong Min \hspace{0.8cm} Minsu Cho \vspace{1.5mm} \\
Pohang University of Science and Technology (POSTECH), South Korea \vspace{1.5mm}\\
\small
\href{http://cvlab.postech.ac.kr/research/HCCNet}{\url{http://cvlab.postech.ac.kr/research/HCCNet}}
}
\maketitle

\begin{figure*}[h]
    \begin{center}
        \includegraphics[width=0.99\textwidth]{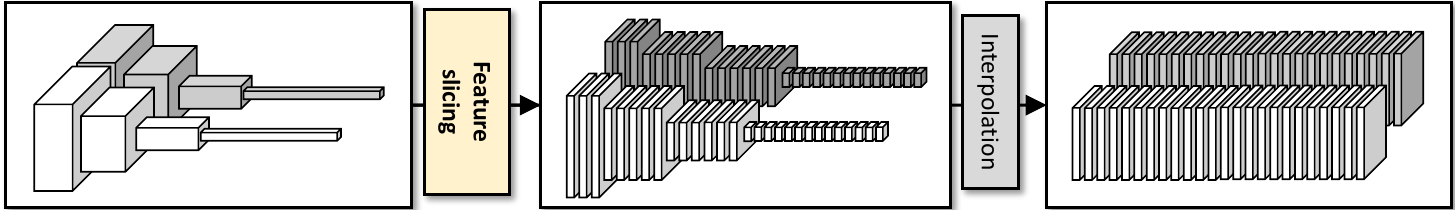}
    \end{center}
    \vspace{-5.0mm}
    \caption{\textbf{Visualization of feature slicing.} 
        Feature slicing slices the multi-level features with varying channel dimensions to an increased number of equi-channel features for the computation of rich hypercolumn correlation.
        }
        \label{fig:supp_wacv_fragmentation_viz}
\end{figure*}

In this supplementary material, we provide additional implementation details (Section~\ref{sec:supp_implementation_details}), quantitative results with analyses (Section~\ref{sec:supp_quantitative}), and additional qualitative results (Section~\ref{sec:supp_qualitative}) of our proposed \methodName, an efficient and effective semantic matching learner on a hypercolumn correlation.

\section{Additional implementation details.}
\label{sec:supp_implementation_details}

\smallbreak
\noindent
\textbf{Visualization of feature slicing.}
The visualization of our proposed feature slicing is provided in Figure \ref{fig:supp_wacv_fragmentation_viz}.
Given $L$ intermediate feature maps whose channel dimensions sum up to $C$, using the slice size of $N$ results in a total of $G=\frac{C}{N}$ feature slices \textit{with the same channel dimensions} for hypercolumn correlation computation.

\smallbreak
\noindent
\textbf{Hyperparameters for flow field formation and keypoint transfer.}
We provide detailed hyperparameters for our methods explained in the paper.
$\mathbf{G}^\mathbf{p} \in \mathbb{R}^{30 \times 30}$, the 2-dimensional Gaussian kernel centered on $\mathbf{p} = \text{arg max}_{\mathbf{y}} \mathbf{C}^{\text{out}}_{\mathbf{x}, \mathbf{y}}$ in Eqn. 8 of the main paper, has a standard deviation of 10.
The distance threshold $\tau$ for the soft sampler in Eqn. 9 of the main paper is set to 0.1 during training, and 0.05 during inference.

\smallbreak
\noindent
\textbf{Coordinate normalization of dense flow field.}
Adhering to the conventions used in ~\cite{lee2019sfnet, min2021chm, swkim2022tfmatcher}, we normalize the coordinates of the dense regular grids - $\mathbf{P}_{X}, \mathbf{P}_{Y} \in \mathbb{R}^{\bar{H}\bar{W} \times 2}$ for the source and target images, respectively - such that the dense regular grids and the dense flow field will have coordinates in the range $\left[\begin{bmatrix} -1 \\ -1 \end{bmatrix}, \begin{bmatrix} 1 \\ 1 \end{bmatrix}\right]$.
This coordinate normalization aims at numerically stabilizing the loss gradients.

\smallbreak
\noindent
\textbf{ResNet-101 feature extractor settings.}
Throughout training, the weights of the ResNet-101~\cite{he2016deep} feature extractor network are frozen up to \texttt{conv3\_x} to prevent overfitting the network to the train set, since all semantic matching benchmarks have a significantly smaller number of images (\~2K) compared to the ImageNet dataset~\cite{deng2009imagenet}.

\smallbreak
\noindent
\textbf{Additional details for feature slicing and hypercolumn correlation construction.}
As mentioned in Section 3 of the main paper, we view each feature map at every layer as a composition of multiple sub-features concatenated along the channel dimension: $\hat{\mathbf{X}}^{(l)} \coloneqq \text{concat}_{g \in G^{(l)}}\left[ \mathbf{X}^{(g)} \right]$ for all $l \in [L]$, where $G^{(l)}$ is the number of slices used to divide feature map $\mathbf{X}^{(l)}$.
Throughout the main paper, we refer to $G^{(l)}$ as (number of) groups, and the resulting channel size of each slice as the slice size.
In our experimental settings, we use the intermediate features extracted from the bottleneck layers of \texttt{conv3\_x}, \texttt{conv4\_x} and \texttt{conv5\_x}, where the number of bottleneck layers are 4, 23 and 3, respectively, and the intermediate features have channel sizes of 512, 1024 and 2048, respectively.
We use a slice size of 256, resulting in $\frac{(4 \times 512) + (23 \times 1024) + (3 \times 2048)}{256} = 124$ number of groups.
Note that the first column of Table 5 of the main table refers to the slice size, while $G$ in Figure 4 of the main paper refers to the group number.
Therefore, in Figure 4 of the main paper, $G = 30$, $G = 62$, $G = 124$ and $G = 248$ corresponds to slice sizes of None (not sliced, bottleneck layer features are used directly), 512, 256 and 128, respectively.

\smallbreak
\noindent
\textbf{Experimental environment.}
All experiments are run on a machine with an Intel Xeon Gold 6242 CPU and an NVIDIA TITAN RTX with 24G VRAM.
\section{Additional quantitative results and analysis.}
\label{sec:supp_quantitative}

\begin{figure*}[t]
    \begin{center}
        \includegraphics[width=1.0\textwidth]{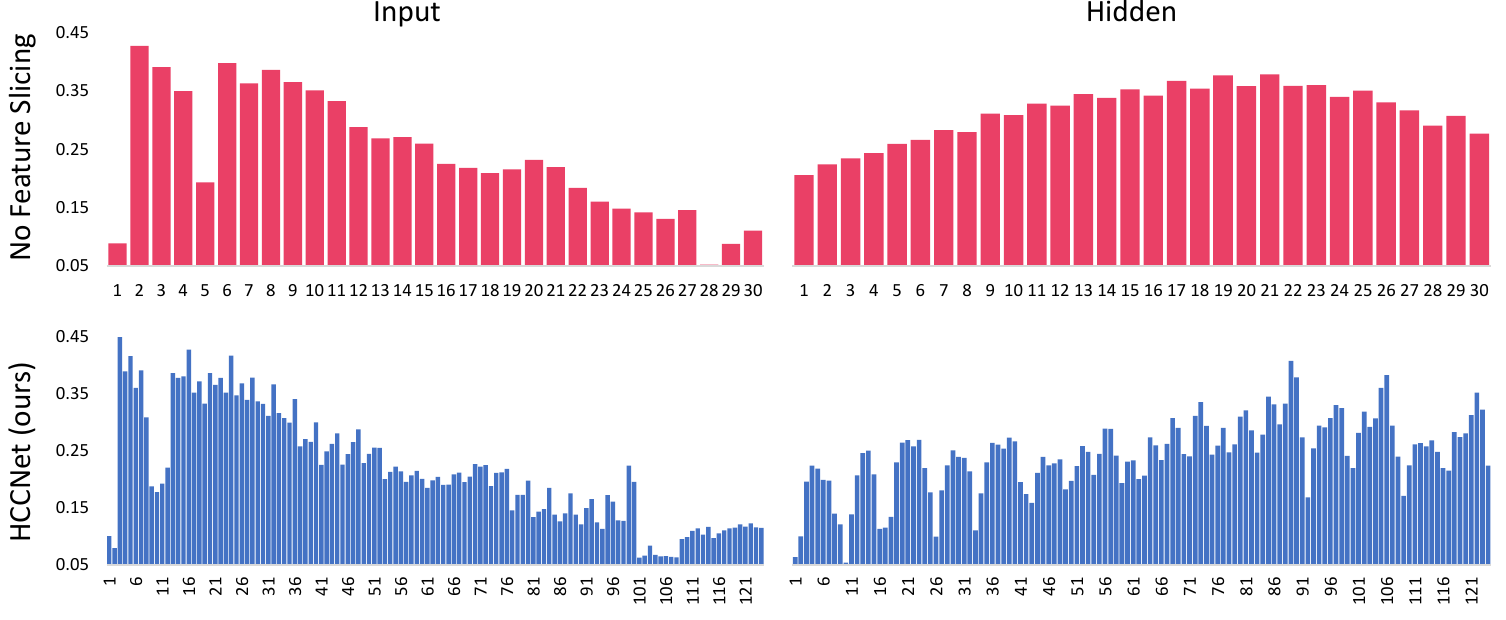}
    \end{center}
    \vspace{-5.0mm}
    \caption{\textbf{Average magnitude for some group $g$ with and without feature slicing.} The high variance of magnitudes when using our feature slicing implies that feature slicing enables fine-grained differentiation of relevant correlation activations to establish more reliable correspondences.}
        \label{fig:feature_frag}
\end{figure*}

\noindent \textbf{Analysis on efficacy of feature slicing.}
\noindent
To provide an in-depth analysis on feature slicing, we plot input $\mathbf{C}$ and hidden $\zeta(\mathbf{C}\mathbf{W}_{\text{hid}})$ correlation statistics from our point-wise convolution in Fig.~\ref{fig:feature_frag} 
where each bar represents average magnitude (y-axis) for some group $g$ (x-axis) of the hypercolumn correlation\footnote{E.g., $\frac{1}{HWHW}\sum_{(\mathbf{x}, \mathbf{y})}|\mathbf{C}_{\mathbf{x}, \mathbf{y}, g}|$ given $\mathbf{C}$ as input.}.
Both models exhibit similar trends, but 
the statistics of using our proposed feature slicing (blue) shows higher variance in magnitudes,  indicating fine-grained differentiation of relevant correlation activations. 
In contrast, the statistics without feature slicing (red) is relatively uniform, implying the inability to exploit rich channel-wise information of the input feature pairs $\{(\mathbf{\hat{X}}^{(l)}, \mathbf{\hat{Y}}^{(l)})\}_{l \in [L]}$.

\smallbreak
\noindent 
\textbf{Influence of image size on \methodName.} 
\begin{table}[h]
    \centering
       \begin{tabular}{cccccc}
                \toprule
                 \multirow{2}{*}{Img size}&  \multicolumn{2}{c}{PF-PASCAL} & \multirow{2}{*}{\shortstack{time\\(\emph{ms})}} & \multirow{2}{*}{\shortstack{mem.\\(GB)}} &\multirow{2}{*}{\shortstack{FLOPs\\(G)}}\\
                                 
                 & 0.05 & 0.1 & & \\
                
                \midrule

                224 & 60.4 & 88.1 & 29 & 2.0 & 1.5 \\ 
                240 (Ours) & 80.2 & 92.4 & 30 & 2.0 & 1.7 \\
                256 & 77.9 & 92.0 & 30 & 2.2 & 2.0 \\
                320 & 71.2 & 89.7 & 38 & 3.1 & 4.0 \\
                384 & 67.1 & 87.3 & 49 & 4.3 & 7.3 \\
                \bottomrule
        \end{tabular}    
        \caption{\textbf{Results of \methodName when using varying image sizes.} The results show that our current setting of using 240 as the height and width of the images yields the best results.}
        \label{tbl:rbt_image_size}
\end{table}
In Table~\ref{tbl:rbt_image_size}, we compare the results of \methodName when using varying image sizes.
It can be seen that our image size of $240\times240$ achieves the best performance-compute tradeoff.
We suggest that the optimal image size depends on the method, as existing methods show optimal performances at varying image dimensions.

\smallbreak
\noindent
\textbf{Analysis on feature slicing strategy.} 
\begin{table}[h] 
    \centering
    \begin{tabular}{ccc}
                \toprule
                 \multirow{2}{*}{Slicing strategy}&  \multicolumn{2}{c}{PF-PASCAL}\\
                                 
                 & 0.05 & 0.1 \\
                
                \midrule

                None & 77.3 & 92.2\\ 
                Neighbour (Ours)  & 80.2 & 92.4  \\
                Random$_\textrm{test}$  & 54.9$_{\pm 0.40}$ & 77.7$_{\pm 0.62}$  \\   
                Even dist. & 75.8 & 91.1  \\
            
                \bottomrule
        \end{tabular}
        \caption{\textbf{Results of \methodName when using varying feature slicing strategies.} It can be seen that our current choice of assigning neighbouring features into a slice shows better performances compared to random or evenly distributed assignment of features.}
        \label{tbl:rbt_fragmentation_strategy}
\end{table}
In Table \ref{tbl:rbt_fragmentation_strategy}, we
compare against other potential slicing strategies.
\methodName clusters neighbouring features into a slice (\textit{Neighbour}).
\textit{Random$_\textrm{test}$} refers to random clustering of features at test time for the \textit{Neighbour} model, without replacement.
\textit{Even dist.} clusters the features in an evenly distributed manner \textit{i.e.,} features of a slice comes from across the entire list of features.
Our strategy yields the best performance.

\begin{table*}
\centering
\scalebox{0.95}{
\begin{tabular*}{\textwidth}{l@{\extracolsep{\fill}}cccccccc}
                \toprule
                \multirow{3}{*}{Method} & \multicolumn{2}{c}{SPair-71k} & \multicolumn{2}{c}{PF-PASCAL} &
                \multirow{3}{*}{Image size} &
                \multirow{3}{*}{\shortstack{time\\(\emph{ms})}} & \multirow{3}{*}{\shortstack{memory\\(GB)}} &\multirow{3}{*}{\shortstack{FLOPs\\(G)}}\\
                
                & \multicolumn{2}{c}{@$\alpha_{\text{bbox}}$} & \multicolumn{2}{c}{@$\alpha_{\text{img}}$} & \\ 
                 
                 & 0.05 & 0.1  & 0.05 & 0.1 & \\
                 
                 \midrule
                 
                 VAT~\cite{hong2022cost}   & \underline{31.7} & 54.2 & 78.3 & 92.3 & $512\times512$ & 127 & 3.6 & 68.0  \\
                 
                 SCorrSAN~\cite{huang2022learning}   & - & 49.8 & - & - & $256\times256$ & \textbf{28} & \textbf{1.5} & \underline{2.1} \\
                 
                 SCorrSAN$_\textrm{MT}$~\cite{huang2022learning}  & - & \textbf{55.3} & \textbf{81.5} & \textbf{93.3} & $256\times256$ & \textbf{28} & \textbf{1.5} & \underline{2.1} \\

                 \midrule
                 
                 \methodName (ours) & \textbf{35.8} & \underline{54.8} & \underline{80.2} & \underline{92.4} & $240 \times 240$ & \underline{30} & \underline{2.0} & \textbf{1.7}  \\
                 
                 \bottomrule
        \end{tabular*}
        }
        \vspace{3.0mm}
        
        \caption{\textbf{PCK results on SPair-71k and PF-PASCAL in comparison to VAT and SCorrSAN.} 
        We compare \methodName with SoTA semantic matching methods of VAT~\cite{hong2022cost} SCorrSAN~\cite{huang2022learning}, where \methodName still exhibits competitive performance and the lowest FLOPs.
        While SCorrSAN$_\textrm{MT}$ shows the best results, their main contributions are distillation and label densification schemes which are both complementary to the model architecture used.
        SCorrSAN shows the results when the label densification and distillation schemes are unused.
        Therefore, \methodName is also expected to benefit significantly when these schemes are applied.
        }
        \label{tbl:supp_eccv}
        \vspace{-1.0mm}
\end{table*}

\smallbreak
\noindent
\textbf{Comparison to SCorrSAN~\cite{huang2022learning}.}
We compare \methodName against SoTA semantic matching methods of VAT~\cite{hong2022cost} and SCorrSAN~\cite{huang2022learning}.
For SCorrSAN, the code was released but not its pretrained weights - we therefore take the PCK results from their paper, and report the latency, FLOPs and memory using an unpretrained version of SCorrSAN.
We include VAT specifically for a comparative evaluation of \methodName on a lower PCK threshold of SPair-71k dataset (\ie at $a_\textrm{bbox} = 0.05$).
Table~\ref{tbl:supp_eccv} presents the results. 

It can be seen that \methodName outperforms VAT on all datasets at all thresholds, while incurring significantly lower latency (approx. 3 times), memory (approx. 1.5 times) and FLOPs (40 times). 
For SCorrSAN, we report two results - one without both knowledge distillation and label densification (SCorrSAN), and the other with both (SCorrSAN$_\textrm{MT}$).
SCorrSAN exhibits notably lower FLOPs compared to \methodName thanks to their efficient spatial context encoder module, albeit incurring higher FLOPs.
While their reported performance is considerably lower that those of VAT or \methodName, they can leverage the lightweightedness of their SCorrsSAN model to perform knowledge distillation and label densification to largely boost their performance (SCorrSAN$_\textrm{MT}$).
We highlight that the knowledge distillation and label densification scheme of SCorrSAN$_\textrm{MT}$ is complementary to the model architecture, and we expect a significant performance improvement once we apply SCorrSAN's training scheme to \methodName.

\smallbreak
\noindent
\textbf{Ablation study and analyses on SPair-71k.}
We conduct the same set of ablative and analytical experiments in the main paper, but on the SPair-71K dataset instead of PF-PASCAL dataset.

\begin{table}[!h]
    \centering
       \begin{tabular*}{0.48\textwidth}{cccccccc}
                \toprule
                \multicolumn{4}{c}{\texttt{conv} used} &  \multicolumn{2}{c}{SPair-71k} & \multirow{3}{*}{\shortstack{mem.\\(GB)}} &\multirow{3}{*}{\shortstack{FLOPs\\(G)}}\\
                
                \multirow{2}{*}{2\_x} & \multirow{2}{*}{3\_x}  & \multirow{2}{*}{4\_x}& \multirow{2}{*}{5\_x} & \multicolumn{2}{c}{@$\alpha_{\text{bbox}}$} & & \\ 
                 
                & & & & 0.05 & 0.1 & & \\
                
                \midrule

                $\times$ & $\times$ & $\times$ & \checkmark & 26.8 & 48.3 & 1.9 & 0.9  \\ 
                
                $\times$ & $\times$ & \checkmark & $\times$ & 32.9 & 51.1 & 1.9 & 1.3 \\

                $\times$ & $\times$ & \checkmark & \checkmark & \textbf{35.8} & 54.3 & 1.9 & 1.6 \\

                $\times$ & \checkmark & \checkmark & \checkmark & \textbf{35.8} & \textbf{54.8} & 2.0 & 1.7 \\

                \checkmark & \checkmark & \checkmark & \checkmark & 34.3 & 53.4 & 2.0 & 1.7 \\
                \bottomrule
        \end{tabular*}
        \vspace{1.5mm}
        \caption{\textbf{Ablation study on the backbone bottleneck features used.}
        The results show that our current setting of using \texttt{conv3\_x} to \texttt{conv5\_x} yields the best results.
        }
        \vspace{-2mm}
        \label{tbl:supp_ablation_conv}
\end{table}

Table~\ref{tbl:supp_ablation_conv} shows that our choice of using the intermediate features extracted from \texttt{conv3\_x} to \texttt{conv5\_x} yields the best results on the SPair-71k dataset as well, which is consistent with the results on the PF-PASCAL dataset (Table 4 of the main paper).

\begin{table}[!h]
    \centering
       \begin{tabular*}{0.48\textwidth}{c@{\extracolsep{\fill}}ccccc}
                \toprule
                 \multirow{3}{*}{Slice size}&  \multicolumn{2}{c}{SPair-71k} & \multirow{3}{*}{\shortstack{time\\(\emph{ms})}} & \multirow{3}{*}{\shortstack{mem.\\(GB)}} &\multirow{3}{*}{\shortstack{FLOPs\\(G)}}\\
                
                & \multicolumn{2}{c}{@$\alpha_{\text{bbox}}$} & & & \\ 
                 
                 & 0.05 & 0.1 & & \\
                
                \midrule

                - & 34.4 & 53.9 & 20 & 2.0 & 0.9 \\ 
                512 & 35.3 & 54.5 & 24 & 2.0 & 1.1 \\
                256 (Ours) & \textbf{35.8} & \textbf{54.8} & 30 & 2.0 & 1.7 \\
                128 & 34.7 & 52.9 & 43 & 2.2 & 4.0 \\
                64 & 35.1 & 52.8 & 70 & 2.2 & 13.3 \\
                32 & 33.3 & 52.5 & 127 & 2.7 & 50.7 \\
            
                \bottomrule
        \end{tabular*}
        \vspace{2.0mm}
        \caption{\textbf{Ablation study on the slice size used on the SPair-71k dataset.}
        The results show that our current setting of using the chunk size of 256 yields the best trade-off between performance and efficiency. 
        }
        \label{tbl:supp_ablation_chunk}
\end{table}

Table~\ref{tbl:supp_ablation_chunk} shows that our choice of using a slice size of 256 strikes the best balance between performance and efficiency, which is also consistent with the results on PF-PASCAL from the main paper (Table 5 of the main paper).

\begin{table}[!h]
    \centering
       \begin{tabular*}{0.48\textwidth}{c@{\extracolsep{\fill}}cc}
                \toprule
                 \multirow{3}{*}{Activation function}&  \multicolumn{2}{c}{SPair-71k} \\ 
                
                & \multicolumn{2}{c}{@$\alpha_{\text{bbox}}$}\\ 
                 
                 & 0.05 & 0.1 \\
                
                \midrule

                ReLU & \textbf{35.8} & 54.4 \\ 
                Sigmoid & 35.4 & 54.0 \\
                Tanh & \textbf{35.}8 & \textbf{54.8} \\
            
                \bottomrule
        \end{tabular*}
        \vspace{2.0mm}
        \caption{\textbf{Ablation study on the non-linear activation function used.}
        Using the Tanh activation function yields the best results, over ReLU or Sigmoid activation functions.
        }
        \label{tbl:supp_ablation_nonlinearity}
        \vspace{-4.0mm}
\end{table}

Table~\ref{tbl:supp_ablation_nonlinearity} shows that our choice of Tanh yields the best results on the SPair-71k dataset compared to using ReLU or Sigmoid, which is consistent with the results on the PF-PASCAL dataset (Table 6 of the main paper).

\smallbreak
\noindent
\textbf{Results when using HPF-selected layers for hypercolumn correlation construction.}
HPF~\cite{min2019hyperpixel} propose to represent images using hyperpixels that leverage a small number of relevant features selected among early to late layers of the backbone feature extractor.
Table~\ref{tbl:supp_hpf} shows that using all bottleneck layers from \texttt{conv3\_x} to \texttt{conv5\_x} yields significantly better results compared to using HPF~\cite{min2019hyperpixel}-selected bottleneck layers for each dataset, substantiating our design choice over using HPF-selected bottleneck layers.

\begin{table}
\centering{
\begin{tabular*}{0.48\textwidth}{l@{\extracolsep{\fill}}cccc}
                \toprule
                \multirow{3}{*}{Method} & \multicolumn{2}{c}{SPair-71k} & \multicolumn{2}{c}{PF-PASCAL} \\

                & \multicolumn{2}{c}{@$\alpha_{\text{bbox}}$} & \multicolumn{2}{c}{@$\alpha_{\text{img}}$} \\ 
                 
                 & 0.05 & 0.1  & 0.05 & 0.1 \\
                 
                 \midrule
                 
                 \methodName$_\textrm{HPF}$ & 31.7 & 49.7 & 76.3 & 90.8  \\
                 
                 \methodName (ours) & \textbf{35.8} & \textbf{54.8} & \textbf{80.2} & \textbf{92.4}  \\
                 
                 \bottomrule
        \end{tabular*}
        }
        
        \caption{\textbf{PCK results on SPair-71k and PF-PASCAL when using bottleneck layers selected by HPF only vs. ours.} 
        The results show that using all bottleneck layers from \texttt{conv3\_x} to \texttt{conv5\_x} yields significantly better results compared to using HPF~\cite{min2019hyperpixel}-selected bottleneck layers for each dataset.
        }
        \label{tbl:supp_hpf}
        \vspace{-1.0mm}
\end{table}

\begin{figure}[t]
    \begin{center}
    \includegraphics[width=1.0\linewidth]{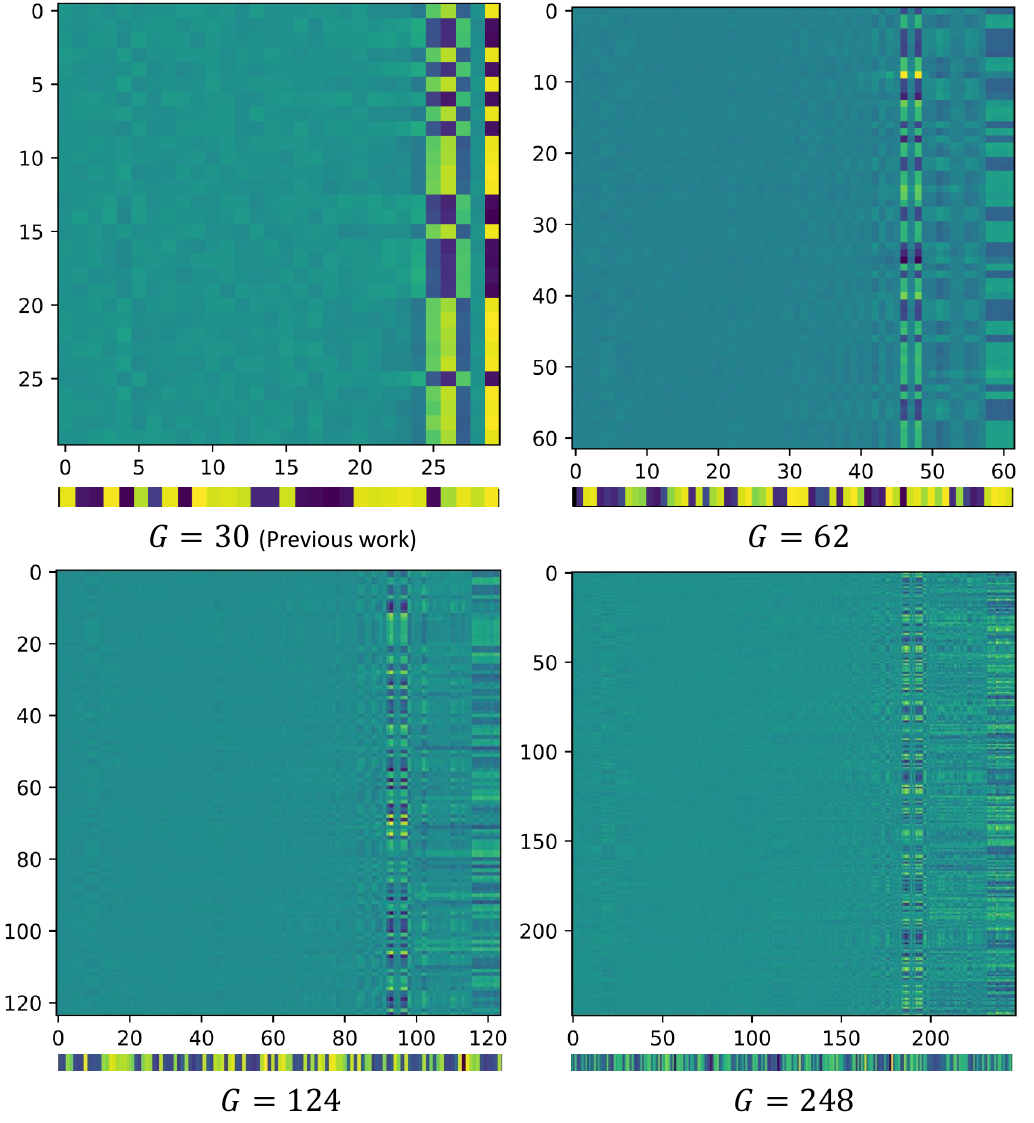}
    \end{center}
     \vspace{-2.5mm}
    \caption{Visualization of learned weight matrices of $\mathbf{W}_{\text{hid}} \in \mathbb{R}^{G \times D_{\text{hid}}}$ (top) and $\mathbf{W}_{\text{out}} \in \mathbb{R}^{D_{\text{hid}} \times 1}$ (bottom) under varying $G = D_{\text{hid}} \in \{30, 62, 124, 248\}$.
    Trained on SPair-71k.}
    \vspace{-3.0mm}
    \label{fig:supp_feature_densification}
\end{figure}

\smallbreak
\noindent
\textbf{Additional feature slicing analysis on SPair-71k.} 
To further investigate the impact of channel aggregation on the hypercolumn correlation, we visualize learned weight matrices $\mathbf{W}_{\text{hid}}$ and $\mathbf{W}_{\text{out}}$ with four different groups denoted by $G \in \{30, 62, 124, 248\}$ in Fig.~\ref{fig:supp_feature_densification} when trained on the SPair-71k dataset, as opposed to Figure 4 of the main paper which illustrates the analysis of feature slicing when trained on the PF-PASCAL dataset.
We consistently observe that the weight magnitudes are significantly higher (in yellow) at deeper layers, particularly at \texttt{conv4\_x} and \texttt{conv5\_x}.
As we increase the number of groups utilized for feature slicing (\ie decrease the feature slice size), we find that the network carries out {\em fine-grained channel selection}, as evidenced by the weight visualization of $\mathbf{W}_{\text{hid}}$, verifying the efficacy of performing position-wise channel aggregation on hypercolumn correlation using diverse visual cues.
Compared to the weight magnitudes of $\mathbf{W}_{\text{hid}}$ that are focused on specific groups, those of $\mathbf{W}_{\text{out}}$ are relatively evenly dispersed in order to effectively aggregate the information from the first channel aggregation to provide a reliable refined correlation map.
The observations are overall consistent across the two datasets, demonstrating the efficacy of \methodName regardless of the dataset it is trained on.

\smallbreak
\noindent \textbf{Variance of multiple runs.}

\begin{table}[h]
    \centering
       \begin{tabular}{ccc}
                \toprule
                 \multirow{2}{*}{Slice size}&  \multicolumn{2}{c}{PF-PASCAL @ $\alpha_{\text{img}}$}\\
                                 
                 & 0.05 & 0.1 \\
                
                \midrule
                512  & 77.8$_{\pm 1.16}$ & 91.9$_{\pm 0.18}$  \\   
                256 (Ours)  & 80.0$_{\pm 0.49}$ & 92.2$_{\pm 0.27}$  \\
                128 & 79.6$_{\pm 0.64}$ & 92.0$_{\pm 0.21}$\\ 
           
                \bottomrule
        \end{tabular}
        \caption{\textbf{Results of \methodName with varying slice size over multiple runs.} Using the slice size of 256 yields consistently better results over other sizes.}
        \label{tbl:rbt_fragmentation_size_variance}
\end{table}

\noindent
In Table \ref{tbl:rbt_fragmentation_size_variance}, we report the PCK variance of our model over three runs with different random seed, while varying the slice size for our feature slicing scheme.
The slice size of 256 still shows to be optimal, further verifying our design choice of \methodName.

\section{Additional qualitative results and analysis.}
\label{sec:supp_qualitative}

\begin{figure*}
    \begin{center}
        \includegraphics[width=\linewidth]{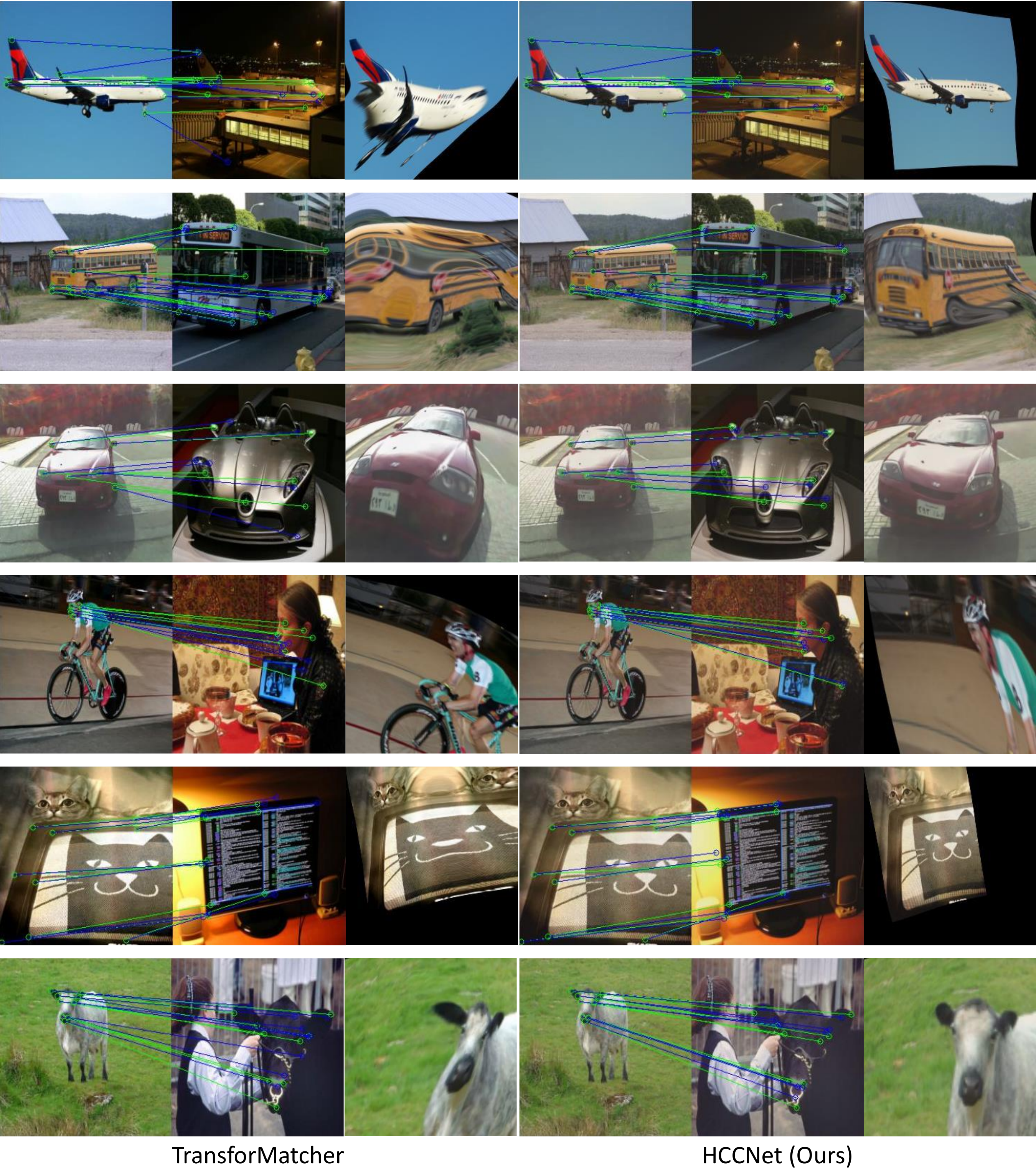}
    \end{center}
    \vspace{-6mm}
      \caption{\textbf{Qualitative comparison of \methodName against TransforMatcher}~\cite{swkim2022tfmatcher}. Green lines represent ground truth correspondences, and blue lines represent predicted correspondences. The source images are TPS warped to the target image using the predicted correspondences for better comparison and visibility. Best viewed in electronics.
}
\label{fig:supp_qual_1}
\end{figure*}

In Figure~\ref{fig:supp_qual_1} we qualitatively compare \methodName with TransforMatcher~\cite{swkim2022tfmatcher} on the SPair-71k dataset, where it can be seen that \methodName established more reliable and accurate correspondences in comparison.
Also, we further compare \methodName with TransforMatcher on image pairs with larger variations in viewpoint/occlusion/truncation from the SPair-71k dataset, where \methodName shows stronger robustness and reliability.
Note that we reproduced TransforMatcher using their released code to obtain these qualitative results.
To enhance the visibility of the qualitative results for better comparison, the source images are TPS-transformed~\cite{donato2002approximate} to the target images using the predicted correspondences to align the common instances in each image pair.

\begin{figure*}
    \begin{center}
        \includegraphics[width=\linewidth]{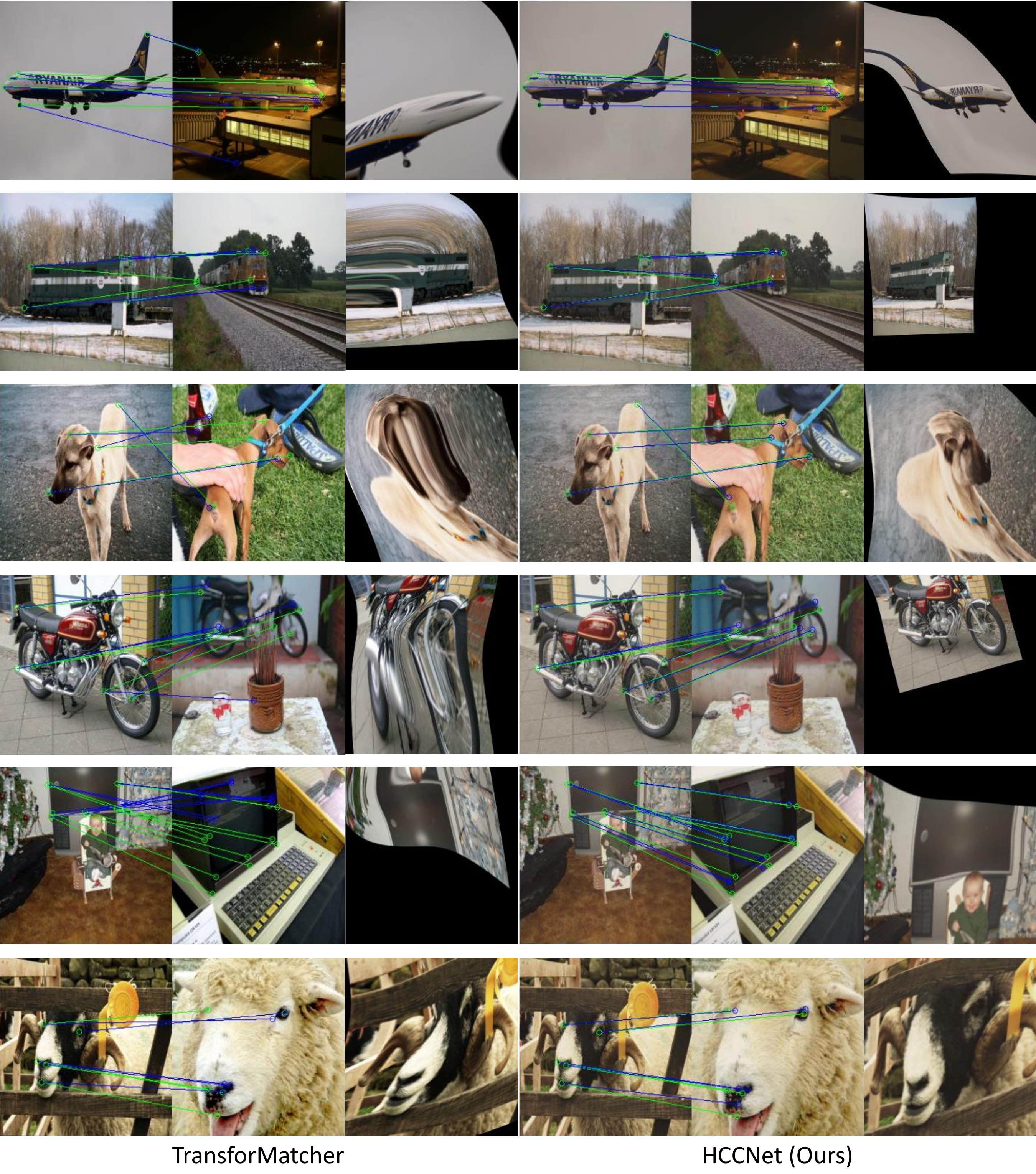}
    \end{center}
    \vspace{-6mm}
      \caption{\textbf{Qualitative comparison of HCCNet against TransforMatcher~\cite{swkim2022tfmatcher} under larger viewpoint/occlusion/truncation variations.} Green lines represent ground truth correspondences, and blue lines represent predicted correspondences. The source images are TPS warped to the target image using the predicted correspondences for better comparison and visibility. Best viewed in electronics.
}
\label{fig:supp_qual_2}
\end{figure*}

\end{document}